# The DLR Hierarchy of Approximate Inference

**Michal Rosen-Zvi**
Computer Sciences and Engineering
Hebrew University of Jerusalem
Jerusalem, Israel 91904

**Michael I. Jordan**
Computer Science and Statistics
University of California
Berkeley, CA 94720

**Alan L. Yuille**
Statistics and Psychology
University of California
Los Angeles, CA 90095

## Abstract

We propose a hierarchy for approximate inference based on the Dobrushin, Lanford, Ruelle (DLR) equations. This hierarchy includes existing algorithms, such as belief propagation, and also motivates novel algorithms such as factorized neighbors (FN) algorithms and variants of mean field (MF) algorithms. In particular, we show that extrema of the Bethe free energy correspond to approximate solutions of the DLR equations. In addition, we demonstrate a close connection between these approximate algorithms and Gibbs sampling. Finally, we compare and contrast various of the algorithms in the DLR hierarchy on spin-glass problems. The experiments show that algorithms higher up in the hierarchy give more accurate results when they converge but tend to be less stable.

## 1 INTRODUCTION

The design and analysis of approximate inference algorithms for large-scale models remains one of the central problems in the graphical models field. Much progress has been made in recent years by taking a *variational* point of view—the exact inference problem (e.g., marginalization or maximization) is expressed as an optimization problem, an approximation is made to the optimization functional or the constraint set (or both), and approximate inference algorithms are expressed in terms of the minimization of the perturbed problem. Several algorithms that originally entered the graphical model field as heuristics—including mean field (MF) algorithms and belief propagation (BP)—have been usefully recast within a variational framework (Amit, 1992, Yedidia et al., 2001). The variational framework has also been used to derive a variety of new algorithms.

There are, however, some limitations to the variational point of view. Consider first the (loopy) BP algorithm, one of the most successful approximate inference algorithms. Although BP can be viewed variationally as the minimization of the Bethe free energy (Yedidia et al., 2001), it is not the case that the BP iteration is a descent step in Bethe free energy, and thus its motivation from the variational framework is not entirely straightforward. Moreover, although algorithms that are descent algorithms can be developed, they do not necessarily outperform BP (Yuille, 2002, Welling and Teh, 2001, Kappen and Wiegerinck, 2002, Heskes et al., 2003, Ikeda et al., 2004), a fact which suggests that the behavior of BP may not be entirely understandable from its characterization as a variational algorithm. Second, the characterization of algorithms as variational has thus far not proved very helpful in suggesting links between those algorithms and Markov chain Monte Carlo (MCMC), the other main source of approximate inference algorithms. Such links would be helpful in the design of hybrid algorithms. Finally, although the variational framework naturally suggests certain kinds of approximations, there may be other approximations that are also worth exploring.

This paper presents a framework for the design and analysis of approximate inference algorithms that is complementary to the variational framework. The framework is based on a linear system of equations known as the Dobrushin, Lanford and Ruelle (DLR) equations (Georgii, 1988, Parisi, 1988). Solving these equations exactly is tantamount to performing exact inference, a task that is deemed impossible for the purposes of this paper. Instead, we design inference algorithms by choosing subsets of the DLR equations. Special cases of this general approach already exist in the literature (Leisink and Kappen, 2001, Pretti and Pelizzola, 2003), but the framework has not yet been exploited systematically as a source of inference algorithms. In the current paper we show that many existing approximate inference algorithms, including BP, have fixed points corresponding to solutions of reduced forms of the DLR equations. In addition, we prove that all extrema of the Bethe free energy correspond to such solutions; this implies that the descent algorithms mentioned above also fit into this framework. We also show that a novel class of algorithms that we refer to as *factorized neighbor* (FN) algorithms arise naturally from the framework. Finally, we show how the framework allows us to design novel variants of MF algorithms.

We also show that the DLR framework has a natural relationship to MCMC algorithms, specifically to the Gibbs sampler. This relationship has long been known for MF algorithms (Amit, 1992)—the current paper extends it to BP and other algorithms.

The paper is organized as follows. Section 2 presents the DLR equations and describes an approximation hierarchy based on these equations. Section 3 shows how the resulting algorithms can be related to MCMC methods. Section 4 provides a variational interpretation of our results. Section 5 presents numerical results and we present our conclusions in Section 6.

## 2 THE DLR EQUATIONS AND THEIR APPROXIMATIONS

The DLR equations express the dependencies of subsets of nodes on their Markov blanket:

$$P(x_R) = \sum_{x_{N(R)}} P(x_R|x_{N(R)})P(x_{N(R)}), \quad \forall R \in \Lambda, \quad (1)$$

where $R$ is an arbitrary subset of nodes, where $\Lambda$ is the set of all such subsets (i.e., a power set), and where $N(R)$ is the neighborhood (i.e., Markov blanket) of the subset $R$.

The DLR equations are a set of linear equations for the unknowns $P(x_R)$. Solving these equations is impractical because of the cardinality of $\Lambda$; instead, we approximate them as follows.

First, we restrict the power set $\Lambda$ to a subset $\Lambda_A$ and retain only the equations corresponding to subsets $R \in \Lambda_A$. Second, we define new variables $b_R(x_R)$ and $B_{N(R)}(x_{N(R)})$ to replace the marginal probabilities $P(x_R)$ and $P(x_{N(R)})$ on the left-hand side and right-hand side in the DLR equations. Third, we define the variables $\{B_{N(R)}(x_{N(R)})\}$ as functions of the variables $\{b_R(x_R)\}$—different choices of functional dependence yield different approximation algorithms.

This procedure yields a set of equations that we will refer to as the *reduced DLR equations*:

$$b_R(x_R) = \sum_{x_{N(R)}} P(x_R|x_{N(R)})B_{N(R)}(x_{N(R)}), \quad \forall R \in \Lambda_A.$$
(2)

The reduced DLR equations are a generally nonlinear set of equations for the variables $b_R(x_R)$. As we will see, they generally have multiple solutions.

There is also a natural class of iterative update rules whose fixed points are solutions of equation (2). These update rules are of form:

$$b_R^{t+1}(x_R) = \sum_{x_{N(R)}} P(x_R|x_{N(R)})B_{N(R)}^t(x_{N(R)}), \quad \forall R \in \Lambda_A,$$
(3)

where the superscript $t$ labels the iteration.

This framework yields a hierarchy of approximation algorithms, from those that retain only singleton nodes in $\Lambda_A$ to those that retain larger subsets of nodes. In the next three sections we discuss algorithms that arise at several levels of this hierarchy.

### 2.1 LEVEL 1: THE FACTORIZED NEIGHBORS ALGORITHM

In the simplest approximation the set $\Lambda_A$ contains only singletons: $\Lambda_A = \{i\}$, where $i$ ranges over the nodes of the graph. We define variables $b_i(x_i)$ corresponding to the elements of $\Lambda_A$.

We obtain an approximation that we refer to as the *factorized neighbors* (FN) algorithm by defining the neighborhood distribution as the factorized expression $B_{N(i)}(x_{N(i)}) = \prod_{j \in N(i)} b_j(x_j)$.

Given this definition, the reduced DLR equations are satisfied when:

$$b_i(x_i) = \sum_{x_{N(i)}} P(x_i|x_{N(i)})B(x_{N(i)}) , \qquad (4)$$

where here and elsewhere in the paper we drop the subscript on the $B$ variables to avoid cluttering the notation. The corresponding iterative update takes the form:

$$\begin{aligned}
b_i^{t+1}(x_i) &= \sum_{x_{N(i)}} P(x_i|x_{N(i)})B^t(x_{N(i)}) \\
&= \sum_{x_{N(i)}} P(x_i|x_{N(i)}) \prod_{j \in N(i)} b_j^t(x_j) . \quad (5)
\end{aligned}$$

It is clear that fixed points of the update rules (5) are solutions of the reduced DLR equations (4).

Equation (4) has appeared previously in the physics literature as the basis for the so-called "hard spin" mean field equations (see references in Pretti and Pelizzola (2003)).

## 2.2 LEVEL 2: BELIEF PROPAGATION AND ITS VARIANTS

The next level of the hierarchy is to set $\Lambda_A$ to contain singletons and pairs of nodes; formally, $\Lambda_A = \{i, (i,j)\}$. This involves defining variables $b_i(x_i)$ and $b_{ij}(x_i, x_j)$. As in belief propagation, these variables are to be viewed as approximations to marginal probabilities, and we do not require that they are consistent during the evolution of the algorithm.

We define the DLR neighborhood approximations using the Bethe approximations:

$$\begin{aligned} B(x_{N(i)}) &= \sum_{x_i} B(x_i, x_{N(i)}), \\ B(x_{N(i,j)}) &= \sum_{x_i, x_j} B(x_i, x_j, x_{N(i,j)}) , \end{aligned} \quad (6)$$

where

$$\begin{aligned} B(x_i, x_{N(i)}) &= \frac{1}{Z_i} b_i(x_i) \prod_{j \in N(i)} \frac{b_{ij}(x_i, x_j)}{b_i(x_i)} \\ &= \frac{1}{Z_i} \frac{\prod_{j \in N(i)} b_{ij}(x_i, x_j)}{\{b_i(x_i)\}^{n_i - 1}} , \quad (7) \\ B(x_i, x_j, x_{N(i,j)}) &= \frac{1}{Z_{ij}} b_{ij}(x_i, x_j) \prod_{k \in N(i)/j} \frac{b_{ik}(x_i, x_k)}{b_i(x_i)} \\ &\quad \prod_{l \in N(j)/i} \frac{b_{jl}(x_j, x_l)}{b_j(x_j)} . \quad (8) \end{aligned}$$

The quantities $Z_i, Z_{ij}$ are normalization constants. Observe that if the singleton and pairwise distributions are locally consistent (i.e., $\sum_{x_j} b_{ij}(x_i, x_j) = b_i(x_i)$) then the normalization constants become unnecessary and can be set to 1.

These approximations are exact if the graph is a tree, and they can be motivated as a tree-based reparameterization of the joint distribution (Wainwright et al., 2003).

The iterative update equations are of form:

$$\begin{aligned} b_{ij}^{t+1}(x_i, x_j) &= \sum_{x_{N(i,j)}} P(x_i, x_j | x_{N(i,j)}) B^t(x_{N(i,j)}) \\ b_i^{t+1}(x_i) &= \sum_{x_{N(i)}} P(x_i | x_{N(i)}) B^t(x_{N(i)}) . \quad (9) \end{aligned}$$

We now restrict ourselves to distributions $P(x)$ which have only singleton and pairwise potentials, $\{\Psi_i(.)\}$ and $\{\Psi_{ij}(.,.)\}$:

$$P(x) = (1/Z) \prod_i \Psi_i(x_i) \prod_{i,j} \Psi_{ij}(x_i, x_j) . \quad (10)$$

**Theorem 1.** *The update equations (9) for the level 2 approximation correspond to the parallel version of the BP algorithm. The fixed points of the update algorithm (9) correspond to solutions of the reduced DLR equations:*

$$\begin{aligned} b_{ij}(x_i, x_j) &= \sum_{x_{N(i,j)}} P(x_i, x_j | x_{N(i,j)}) B(x_{N(i,j)}) , \\ b_i(x_i) &= \sum_{x_{N(i)}} P(x_i | x_{N(i)}) B(x_{N(i)}) . \quad (11) \end{aligned}$$

Proof. It is known that BP can be expressed as the marginalizations $b_{ij}^{t+1}(x_i, x_j) = \sum_{x_{N(i,j)}} B^t(x_i, x_j, x_{N(i,j)})$ and $b_i^{t+1}(x_i) = \sum_{x_{N(i)}} B^t(x_i, x_{N(i)})$; see Wainwright et al. (2003). In Appendix A we derive this result as a consequence of the fact that $\{b_i(.)\}$ and $\{b_{ij}(.,.)\}$ obey the so-called e-constraint. The e-constraint also implies (see Appendix A) that $B^t(x_i, x_{N(i)}) = P(x_i|x_{N(i)}) B^t(x_{N(i)})$ and $B^t(x_i, x_j, x_{N(i,j)}) = P(x_i, x_j|x_{N(i,j)}) B^t(x_{N(i,j)})$. The update rule (9) follows, provided BP is performed in parallel. □

The fixed points of belief propagation are known to correspond to extrema of the Bethe free energy (Yedidia et al., 2001). We now prove that any extremum of the Bethe free energy corresponds to a solution of the reduced DLR equations.

**Theorem 2.** *Extrema of the Bethe free energy correspond to solutions of the level 2 approximation to the DLR equations.*

Proof. The extrema of the Bethe free energy obey the e-constraint and the m-constraint (see Appendix A). The e-constraint implies that $P(x_i|x_{N(i)}) B(x_{N(i)}) = B(x_i, x_{N(i)})$ and $P(x_i, x_j|x_{N(i,j)}) B(x_{N(i,j)}) = B(x_i, x_j, x_{N(i,j)})$. The m-constraint implies that $b_i(x_i) = \sum_{x_{N(i)}} B(x_i, x_{N(i)})$ and $b_{ij}(x_i, x_j) = \sum_{x_{N(i,j)}} B(x_i, x_j, x_{N(i,j)})$. (Because the factors $b_{il}(x_i, x_l)/b_i(x_i)$ become conditional distributions $b(x_l|x_i)$ which obey $\sum_{x_l} b(x_l|x_i) = 1$). The result follows. □

This result implies that all algorithms which converge to extrema of the Bethe free energy (Yuille, 2002, Welling and Teh, 2001, Kappen and Wiegerinck, 2002, Wainwright et al., 2003, Heskes et al., 2003, Ikeda et al., 2004) must also converge to solutions of the reduced DLR equations.

Theorem 2 also implies that there can be many solutions to the reduced DLR equations, because there can be many extrema of the Bethe free energy.

## 2.3 LEVEL 1.5: THE CP AND FN2 ALGORITHMS

We obtain other approximations, intermediate between FN and BP, by setting $\Lambda_A = \{(i,j)\}$ and only representing the pairwise marginals $b_{ij}(x_i, x_j)$ (the singletons $b_i(x_i)$ are obtained by marginalization). We use the update rule from equation (9), $b_{ij}^{t+1}(x_i, x_j) = \sum_{x_{N(i,j)}} P(x_i, x_j | x_{N(i,j)}) B^t(x_{N(i,j)})$.

There are two natural choices for the form of the neighborhood distribution $B(x_{N(i,j)})$.

The first choice is to use the Bethe approximation (8) where the singleton distributions $b_i(x_i)$ are replaced by $\sum_{x_j} b_{ij}(x_i, x_j)$. This rederives the CP algorithm of Pretti and Pelizzola (2003). Pretti and Pelizzola (2003) compare the CP algorithm to BP and CCCP (Yuille, 2002), showing that CP performs well and empirically has good convergence properties.

The second choice is to use the factorized neighbor approximation $B(x_{N(i,j)}) = \prod_{k \in N(i,j)} b_k(x_k)$ where the singleton distributions are obtained by $b_i(x_i) = \sum_{j \in N(i)} \sum_{x_j} b_{ij}(x_i, x_j) / |N(i)|$ and $|N(i)|$ is the number of neighbors of the $i$th node.

We call this algorithm *Second-order Factorized Neighbors* (FN2). The update rule for the pairwise marginals is:

$$b_{ij}^{t+1}(x_i, x_j) = \sum_{x_{N(i,j)}} P(x_i, x_j | x_{N(i,j)}) \prod_{k \in N(i,j)} b_k^t(x_k).$$

In Section 5.2 we report experimental results for this algorithm.

## 2.4 MEAN FIELD ALGORITHMS

Surprisingly, mean field (MF) algorithms have not yet appeared in our hierarchy. Traditionally, MF algorithms would be considered the first level approximation with belief propagation as the second level. Yet our approach has replaced the MF approximation by the FN approximation.

We now show that the MF approximation can be understood as an alternative approximation to the DLR equations.

The standard variational derivation of MF algorithms takes as its point of departure the following *free energy* functional:

$$F_{mf}[b] = \sum_x \{\prod_i b_i(x_i)\} \log \frac{\{\prod_i b_i(x_i)\}}{P(x)}. \quad (12)$$

**Theorem 3.** *The extrema of the mean field free energy $F_{mf}[b]$ obey the equations:*

$$\log b_j(x_j) = \sum_{x_{/j}} \log P(x_j | x_{N(j)}) \prod_{k \in N(j)} b_k(x_k) + c_j, \quad (13)$$

*where $c_j$ is a normalization constant and where $x_{/j}$ denotes the configuration of all nodes except node $j$.*

Proof. Express $\log \frac{\{\prod_i b_i(x_i)\}}{P(x)}$ as $\log \frac{b_j(x_j)}{P(x_j | x_{N(j)})} + \log \frac{\prod_{k \neq j} b_k(x_k)}{P(x_{/j})}$. Differentiate with respect to $b_j(x_j)$ and the result follows. □

This has a natural update rule:

$$\log b_j^{t+1}(x_j) = \sum_{x_{/j}} \prod_{k \in N(j)} b_k^t(x_k) \log P(x_j | x_{N(j)}) + c_j. \quad (14)$$

Theorem 3 shows that MF gives an alternative approximation to the DLR equations. But it differs from our previous approximations by containing logarithms.

This motivates alternative MF methods to approximate the DLR equations. For simplicity, we will present them for the Ising spin model:

$$P_I(x) = (1/Z) e^{\sum_{ij} \theta_{ij} x_i x_j + \sum_i \phi_i x_i}, \quad (15)$$

where $x_i \in \{0, 1\}$. For this model, we have the following result:

**Theorem 4.** *The MF algorithm (14) for the Ising spin model $P_I(x)$ can be expressed in the form:*

$$b_i^{t+1}(x_i) = P_I\left(x_i \Big| \sum_{x_{N(i)}} x_{N(i)} B^t(x_{N(i)})\right), \quad (16)$$

*with $B^t(x_{N(i)}) = \prod_{j \in N(i)} b_j^t(x_j)$.*

Proof. The conditional distribution for the Ising model is

$$P_I(x_j | x_{N(j)}) = e^{\phi_j x_j + x_j \sum_i \theta_{ij} x_i} / Z[\{x_i : i \neq j\}]. \quad (17)$$

Substituting into the MF update equation (14) gives the standard MF update rule for the Ising spin model:

$$b_j^{t+1}(x_j) = \frac{e^{\phi_j x_j + \sum_i \theta_{ij} x_j \sum_{\hat{x}_i} \hat{x}_i b_i^t(\hat{x}_i)}}{\sum_{\hat{x}_j} e^{\phi_j \hat{x}_j + \sum_i \theta_{ij} \hat{x}_j \sum_{\hat{x}_i} \hat{x}_i b_i^t(\hat{x}_i)}}. \quad (18)$$

This equation can be readily re-expressed as equation (16). □

The presentation of the MF algorithm as an approximation to the DLR equations for singletons motivates a *second-order MF* (MF2) algorithm for the Ising model by analogy to equation (16). We set

$$b_i^{t+1}(x_i) = \sum_{x_j} P(x_i, x_j | \sum_{x_{N(i,j)}} x_{N(i,j)} B^t(x_{N(i,j)})), \quad (19)$$

where $B(x_{N(i,j)})$ is the factorized neighbors distribution $\prod_{k \in N(i)/j} b_k(x_k) \prod_{l \in N(j)/i} b_l(x_l)$.

This can be re-expressed as

$$\tilde{b}_i^{t+1} = \frac{1}{|N(i)|} \sum_{k \in N(i)} \frac{1}{Z_{ik}^t} \times \sum_{x_k} \exp(\phi_i + \phi_k x_k + \theta_{ik} x_k + \sum_{j \in N(i,k)} \theta_{ij} \tilde{b}_j^t) , \quad (20)$$

where $\tilde{b}_i = b_i(X_i = 1)$ and $Z_{ik}^t = \sum_{x_i, x_k} \exp(\phi_i x_i + \phi_k x_k + \theta_{ik} x_i x_k + \sum_{j \in N(i,k)} \theta_{ij} \tilde{b}_j^t)$.

## 3 MCMC, CHAPMAN-KOLMOGOROV, AND THE GIBBS SAMPLER

We now show that the natural update rule (3) for the reduced DLR equations corresponds to a deterministic approximation to Gibbs sampling.

The update rule for a Markov Chain is given by the Chapman-Kolmogorov equations:

$$\mu^{t+1}(x) = \sum_{x'} K(x|x') \mu^t(x'), \quad (21)$$

where $K(x|x')$ is the transition kernel. The kernel is chosen to satisfy the detailed balance conditions

$$K(x|x') P(x') = K(x'|x) P(x). \quad (22)$$

MCMC consists of simulating equation (21) by repeatedly drawing samples from the transition kernel. Provided weak conditions apply, these are guaranteed to converge to samples from the distribution $P(x)$.

The Gibbs sampler is $K_R(x|x') = P(x_R|x'_{N(R)}) \delta_{x_{/r}, x'_{/r}}$. It can be checked that it satisfies the detailed balance equations (22).

Substituting the Gibbs sampler into the Chapman-Kolmogorov equations (21) and marginalizing yields the update equations:

$$\mu^{t+1}(x_R) = \sum_{x'_{N(R)}} P(x_R|x'_{N(R)}) \mu^t(x'_{N(R)}). \quad (23)$$

Observe that these update equations are identical to those obtained by iterating the reduced DLR equations—cf. equation (3)—provided we replace $\mu(x_R)$ by $b(x_R)$ and $\mu(x_{N(R)})$ by $B(x_{N(R)})$. This gives:

$$b^{t+1}(x_R) = \sum_{x'_{N(R)}} P(x_R|x'_{N(R)}) B^t(x'_{N(R)}) \quad \forall R \in \Lambda_A. \quad (24)$$

This shows that our approximation algorithms can be viewed as a deterministic approximation to the Gibbs sampler. This generalizes a result for MF methods (Amit, 1992). In particular, we obtain:

**Theorem 5.** *The parallel BP update is a deterministic version of the Chapman-Kolmogorov equations using the Gibbs sampler and the Bethe approximations for the neighborhood distributions.*

## 4 VARIATIONAL PRINCIPLES

Although we have motivated the DLR framework as an alternative to the variational framework, the two frameworks are complementary and indeed it is useful to attempt to develop a variational perspective on the algorithms that result from the reduced DLR equations. One particularly natural choice of variational principle to capture solutions of the reduced DLR equations are weighted sums of Kullback-Leibler divergences (WSKL):

$$F_{wskl}[b] = \sum_R \alpha_R \sum_{x_R} b_R(x_R) \log \frac{b_R(x_R)}{\sum_{x_{N(R)}} P(x_R|x_{N(R)}) B(x_{N(R)})}. \quad (25)$$

It is straightforward to show that the extrema of the WSKL (25) correspond to the fixed points of our update algorithms—provided the $\{\alpha_R\}$ are positive. This is because fixed points $b_R^*(x_R)$ of our algorithms satisfy the reduced DLR equations, $b_R^*(x_R) = \sum_{x_{N(R)}} P(x_R|x_{N(R)}) B^*(x_{N(R)})$, and hence $F_{wskl}[b^*] = 0$. This is the smallest value that $F_{wlk}$ can take (provided the $\{\alpha_R\}$ are positive) and so is an extremum of $F_{wlk}$.

Although this suggests that the weights $\{\alpha_R\}$ should be chosen to be positive, by analogy to the Bethe and Kikuchi free energies it may also be useful to allow some of the weights to be negative. In fact we can also show that fixed points of our update equations correspond to extrema of $F_{wskl}$ even when the weights can take negative values.

## 5 EMPIRICAL RESULTS

Many of the algorithms that we have discussed have already been the subject of extensive numerical comparisons. In this section we present a brief selection of additional empirical results that help to situate the novel algorithms that we have discussed.

## 5.1 PHASE TRANSITIONS

One way to compare the different approximations is by using them to predict the critical temperature for the celebrated two-dimensional Ising Spin Model (Parisi, 1988). The model can be thought of an infinite square grid with binary random variables $s_i = \pm 1$ and joint distribution $\exp\left(\sum_{(ij)} J_{ij} s_i s_j / t\right)/Z$. There is a critical temperature above which the averaged spin is zero and below which there is a spontaneous magnetization. The critical temperature can be found by studying the fixed point equations assuming spatial homogeneity: $J_{ij} = 1$ for all neighbors and $J_{ij} = 0$ otherwise. This model has been exactly solved and the critical temperature is known to be $t_c = 2.269$. The critical temperature obtained by the MF approximation is much higher: $t_{MF} = 4$, the critical temperature obtained by the FN/BP is $t_{FN} = 3.089/t_{BP} = 2.885$ (see Parisi, 1988). We solve the fixed point equations for the two novel approximations and obtain $t_{FN2} = 3.025$ and $t_{MF2} = 3.776$. This shows that the FN is a significant improvement over MF2, which is a small improvement compared to MF. Similarly FN2 improves over FN.

## 5.2 NUMERICAL RESULTS

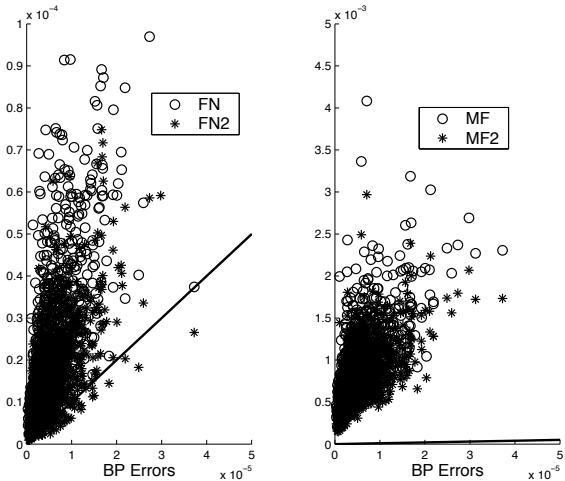

Figure 1: The left panel plots the L1 errors of FN (circles) and FN2 (stars) against the errors for BP, for different realizations of the parameters of an Ising model. The right panel shows the corresponding results for MF (circles) and MF2 (stars).

In this section we present numerical results for approximate inference for the Ising model as defined in equation (15). We experiment with FN, FN2, MF, MF2 and BP on a two-dimensional grid with periodic boundary conditions (i.e., a torus). The interaction weights, $\theta_{ij}$, are drawn from a Gaussian distribution with zero mean. The singleton weights $\phi_i$ are also drawn from a zero-mean Gaussian and are then shifted so that the average of the singleton marginals is $\bar{b}_i = 1/2$. We allow the algorithms to iterate until convergence, assessing convergence by comparing the difference between marginals in consecutive iterations and stopping when this value is less than value $10^{-6}$. (In any case we stop the algorithm when $10^6$ iterations have been performed.)

We start with experiments in a regime that we call the "easy regime," in which BP converges for almost all random realization of the parameters, and in which the resulting estimates are good estimates of the true marginals. Specifically, both the interaction weights and the singleton weights are drawn from a Gaussian with variance equal to 0.1. We used a $4 \times 4$ grid. In Figure 1 we compare results from 1000 different values of the interactions. The left panel plots the $L1$ errors of the singleton marginals from FN (circles) and FN2 (stars) against the BP errors. The right panel shows the $L1$ errors of the singleton marginals from MF (circles) and MF2 (stars) compared to those of BP. The averaged BP error is $5.6 \cdot 10^{-6} \pm 4.3 \cdot 10^{-6}$, the FN error is $2.5 \cdot 10^{-5} \pm 1.9 \cdot 10^{-5}$, the FN2 error is $1.4 \cdot 10^{-5} \pm 1.1 \cdot 10^{-5}$, the MF error is $9.6 \cdot 10^{-4} \pm 4.4 \cdot 10^{-4}$, and the MF2 error is $7.0 \cdot 10^{-4} \pm 3.1 \cdot 10^{-4}$. In general, in this regime, BP provides more accurate results than the other algorithms, although for some realizations the accuracy of FN2 surpasses that of BP. The FN and FN2 algorithms are more accurate than the MF and MF2 algorithms.

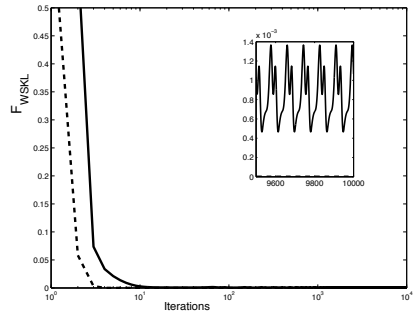

Figure 2: A semi-log plot of the evolution of $F_{wskl}$ for two different realizations of the parameters. The plot shows a typical case in which FN converges (dashed line) and a typical case in which FN fails to converges (solid line). The inset is a zoom of the plot at large values of the iteration number, where it can be verified that the dashed line converges and the solid line does not.

We conducted a second set of experiment in a "hard regime" in which BP fails to converge for most realizations of the parameters. This was achieved by setting the variances of the interaction weights to 4 while retaining the variances of the singleton weights at 0.1. Out of 1000 random choices of parameters BP fails to converge in 712 runs. When it does converge (in 288 runs), the average error of BP is $0.16 \pm 0.16$. FN fails to converge in only 5 of the runs. Moreover, we observed that even in the runs in which FN fails to converge the approximate marginals oscillate around a fixed value, and these marginals can still be useful. In figure 2 we present a typical non-convergent run of FN (solid line) and a typical run where FN does converge (dashed line). The figure plots the evolution of the objective function, $F_{wskl}$, with $\alpha_R = 1$ for $R = \{i\}$ and $\alpha_R = 0$ otherwise. At convergence the objective function should vanish. The inset shows that in the case in which the algorithm does not converge the function $F_{wskl}$ oscillates at a finite (small) distance from zero (solid line), and when the algorithm converges, the objective function vanishes (dashed line).

The average error of FN on these 5 non-converging cases is $0.29 \pm 0.06$. On the convergent cases, the FN error is $0.32 \pm 0.07$, the FN2 error is $0.28 \pm 0.09$, the MF error is $0.41 \pm 0.05$ and the MF2 error is $0.40 \pm 0.05$. Again, we observe that FN2 improves on FN, and MF2 improves (by a smaller amount) on MF.

## 6 CONCLUSIONS

We have presented a unifying framework for the design and analysis of approximate inference algorithms. The framework is based on choosing subsets of the DLR equations. Many existing algorithms fall within this framework—including all those that seek to minimize the Bethe free energy—and novel algorithms can be obtained. We performed computer simulations to further flesh out the DLR hierarchy. For an easy problem, the empirical results improve as we move up the hierarchy; in particular, BP converges reliably and yields accurate results. For a harder problem, the sweet spot appears to be lower down in the hierarchy—BP fails to converge in the majority of the runs and the FN algorithms provide the most satisfactory performance.

In addition, we showed that the DLR framework has a simple and elegant connection with the Gibbs sampler. This may give insight into the relationships between deterministic methods and sampling methods and may help motivate hybrid algorithms.

**Acknowledgements** We thank Max Welling and Bert Kappen for useful discussions. We also wish to thank Intel Corporation and Microsoft Research for their support. Some of this work was done when A.L. Yuille was consulting at Microsoft Research Cambridge.

## 7 APPENDIX A

This appendix presents the properties of BP and the Bethe free energy which are needed to prove Theorems 1 and 2.

### 7.1 BASIC BP AND THE E-CONSTRAINT

The BP update rule, for passing a message from $i$ to $j$, is given by:

$$m_{ij}^{t+1}(x_j) = \sum_{x_i} \Psi_{ij}(x_i, x_j) \Psi_i(x_i) \prod_{k \neq j} m_{ki}^t(x_i). \quad (26)$$

This update equation (26) implies that the singleton and pairwise beliefs are of the following form:

$$b_i^t(x_i) \propto \Psi_i(x_i) \prod_k m_{ki}^t(x_i), \quad (27)$$

$$b_{kj}^t(x_k, x_j) \propto \Psi_k(x_k) \Psi_j(x_j) \Psi_{kj}(x_k, x_j)$$
$$\times \prod_{\tau \neq j} m_{\tau k}^t(x_k) \prod_{l \neq k} m_{lj}^t(x_j). \quad (28)$$

The form defined by equation (28) implies that there is a linear relationship, *the e-constraint*, between the logarithms of the singleton and pairwise beliefs.

This can be seen more easily by re-expressing the beliefs as:

$$b_i^t(x_i) \propto \Psi_i(x_i) e^{\theta_i^t(x_i)}, \quad (29)$$
$$b_{ij}^t(x_i, x_j) \propto \Psi_i(x_i) \Psi_j(x_j) \Psi_{ij}(x_i, x_j) e^{\lambda_{ij}^t(x_j) + \lambda_{ji}^t(x_i)} \quad (30)$$

where the variables $\{\theta_i(x_i)\}$, $\{\lambda_{ij}(x_j)\}$ are related to the messages $\{m_{ij}(x_j)\}$, and to each other, as follows:

$$\lambda_{ji}(x_i) = \sum_{k \in N(i)/j} \log m_{ki}(x_i),$$
$$\theta_i(x_i) = \sum_{k \in N(i)} \log m_{ki}(x_i)$$
$$= \frac{1}{|N(i)| - 1} \sum_{j \in N(i)} \lambda_{ji}(x_i). \quad (31)$$

The e-constraint is equivalent to the requirement (Wainwright et al., 2003) that the beliefs give a re-parameterization of the probability distribution so that:

$$P(x) = \frac{\prod_{ij} \Psi_i(x_i) \Psi_j(x_j) \Psi_{ij}(x_i, x_j)}{\prod_i \{\Psi_i(x_i)\}^{|N(i)|-1}},$$
$$\propto \frac{\prod_{ij} b_{ij}(x_i, x_j)}{\prod_i \{b_i(x_i)\}^{|N(i)|-1}}. \quad (32)$$

It can be shown, using the e-constraints, that the BP update rules (26) can be re-expressed as marginalization of the beliefs (Wainwright et al., 2003). $b_{ij}^{t+1}(x_i, x_j) = \sum_{x_{N(i,j)}} B^t(x_i, x_j, x_{N(i,j)})$ and $b_i^{t+1}(x_i) = \sum_{x_{N(i)}} B^t(x_i, x_{N(i)})$.

## 7.2 BP AND GIBBS SAMPLER

To complete the proof of Theorem 2, we now show that $B^t(x_i, x_{N(i)}) = P(x_i|x_{N(i)})B^t(x_{N(i)})$ and $B^t(x_i, x_j, x_{N(i,j)}) = P(x_i, x_j|x_{N(i,j)})B^t(x_{N(i,j)})$. We require that the beliefs satisfy the e-constraint.

To derive the result for the singleton, we express the conditional distribution $P(x_i|x_{N(i)})$ as

$$\frac{P(x_i, x_{N(i)})}{\sum_{x_i'} P(x_i', x_{N(i)})} = \frac{\Psi_i(x_i) \prod_{j \in N(i)} \Psi_j(x_j)\Psi_{ij}(x_i, x_j)}{\sum_{x_i'} \Psi_i(x_i') \prod_{j \in N(i)} \Psi_j(x_j)\Psi_{ij}(x_i', x_j)}.$$

The form of equations (29,30), together with the relationships in equation (31), *show that this is identical to* $B(x_i|x_{N(i)}) = B(x_i, x_{N(i)})/\sum_{x_i'} B(x_i', x_{N(i)})$, where $B(x_i, x_{N(i)})$ is given by the Bethe approximation (7). We see that $B(x_i, x_{N(i)})$ is proportional to:

$$B(x_i, x_{N(i)}) \propto \Psi_i(x_i)$$
$$\times \prod_{j \in N(i)} \Psi_j(x_j)\Psi_{ij}(x_i, x_j)e^{\lambda_{ji}(x_i)+\lambda_{ij}(x_j)}e^{-(|N_i|-1)\theta_i(x_i)},$$

which can be re-expressed as:

$$\Psi_i(x_i) \prod_{j \in N(i)} \Psi_j(x_j)\Psi_{ij}(x_i, x_j)e^{\lambda_{ij}(x_j)},$$

where we used $\theta_i(x_i) = 1/(N_i - 1)\sum_{j \in N(i)} \lambda_{ji}(x_i)$, equation (31), to cancel $\theta(x_i)$ with the $\{\lambda_{ji}(x_i)\}$. The result for the singleton follows. For the pairwise terms, the true marginals are given by:

$$P(x_i, x_j|x_{N(i,j)}) = \frac{P(x_i, x_j, x_{N(i,j)})}{\sum_{x_i', x_j'} P(x_i', x_j', x_{N(i,j)})}.$$

Using equation (8) for $B(x_i, x_j, x_{N(i,j)})$ and equation (31) to cancel $\theta$ and $\lambda$ terms, we get:

$$B(x_i, x_j, x_{N(i,j)}) \propto \Psi_i(x_i)\Psi_j(x_j)\Psi_{ij}(x_i, x_j)$$
$$\times \prod_{k \in N(i)/j} \Psi_k(x_k)\Psi_{ik}(x_i, x_k)e^{\lambda_{ik}(x_k)}$$
$$\prod_{l \in N(j)/i} \Psi_l(x_l)\Psi_{jl}(x_j, x_l)e^{\lambda_{jl}(x_l)}.$$

Hence

$$\frac{P(x_i, x_j, x_{N(i,j)})}{\sum_{x_i', x_j'} P(x_i', x_j', x_{N(i,j)})} = \frac{B(x_i, x_j, x_{N(i,j)})}{\sum_{x_i', x_j'} B(x_i', x_j', x_{N(i,j)})},$$

because the $\lambda$'s cancel when we compute $B(x_i, x_j|x_{N(i,j)})$.

## 7.3 THE BETHE FREE ENERGY

The Bethe free energy is:

$$F[b] = \sum_{ij} \sum_{x_i, x_j} b_{ij}(x_i, x_j) \log \frac{b_{ij}(x_i, x_j)}{\psi_i(x_i)\psi_j(x_j)\psi_{ij}(x_i, x_j)}$$
$$- \sum_i (n_i - 1) \sum_{x_i} b_i(x_i) \log \frac{b_i(x_i)}{\psi_i(x_i)}$$
$$+ \sum_{i,j} \sum_{x_j} \lambda_{ij}(x_j)\{\sum_{x_i} b_{ij}(x_i, x_j) - b_j(x_j)\}$$
$$+ \sum_{i,j} \sum_{x_i} \lambda_{ji}(x_i)\{\sum_{x_j} b_{ij}(x_i, x_j) - b_i(x_i)\},$$

where the $\lambda_{ij}(x_j), \lambda_{ji}(x_i)$ are Lagrange multipliers used to impose the *m-constraints* $\sum_{x_i} b_{ij}(x_i, x_j) - b_j(x_j)$ and $\sum_{x_j} b_{ij}(x_i, x_j) - b_i(x_i)$.

Extremizing with respect to the beliefs shows that the singleton and pairwise marginals are of form (29,30) and hence satisfy the e-constraint. Imposing the Lagrange multipliers shows that the m-constraints must also be satisfied.